# Improving Prediction Accuracy of Semantic Segmentation Methods Using Convolutional Autoencoder Based Pre-processing Layers

**Hisashi Shimodaira** hshimodaira@hi-ho.ne.jp
No affiliation
2-2-16 Katsuradai, Aoba-Ku, Yokohama-City, 227-0034, Japan

## Abstract

In this paper, we propose a method to improve prediction accuracy of semantic segmentation methods as follows: (1) construct a neural network that has pre-processing layers based on a convolutional autoencoder ahead of a semantic segmentation network, and (2) train the entire network initialized by the weights of the pre-trained autoencoder. We applied this method to the fully convolutional network (FCN) and experimentally compared its prediction accuracy on the cityscapes dataset. The Mean IoU of the proposed target model with the He normal initialization is 18.7% higher than that of FCN with the He normal initialization. In addition, those of the modified models of the target model are significantly higher than that of FCN with the He normal initialization. The accuracy and loss curves during the training showed that these are resulting from the improvement of the generalization ability. All of these results provide strong evidence that the proposed method is significantly effective in improving the prediction accuracy of FCN. The proposed method has the following features: it is comparatively simple, whereas the effect on improving the generalization ability and prediction accuracy of FCN is significant; the increase in the number of parameters by using it is very small, and that in the computation time is substantially large. In principle, the proposed method can be applied to other semantic segmentation methods. For semantic segmentation, at present, there is no effective way to improve the prediction accuracy of existing methods. None have published a method which is the same as or similar to our method and none have used such a method in practice. Therefore, we believe that our method is useful in practice and worthy of being widely known and used.

## 1. INTRODUCTION

Semantic segmentation is a fundamental step in the field of computer vision. It assigns per-pixel predictions of object classes for a given image. Semantic segmentation is a challenging task in two respects: an object associated with a specific concept (semantics) should be recognized and classified correctly (classification), and the classification label for a pixel must be located at the correct position of the object. The segmentation model should simultaneously address these two contradictory issues.

Many computer vision and machine learning techniques for semantic segmentation were proposed before the deep learning era, and most of them relied on hand-engineered features to classify pixels. Recently, the emergence of deep learning techniques for neural networks and convolutional neural networks has led to a revolution in image classification and semantic segmentation methods. In pioneering works, VGG 16-layer net (VGG16 net) [1], fully convolutional network (FCN) [2, 3], U-Net [4], etc., have been proposed. FCN was developed by converting the VGG16 net for image classification to a semantic segmentation network. FCN has paved the way for deep learning -based semantic segmentation. To date, numerous novel deep learning-based methods have been proposed, for example, [5, 6], which showed remarkable improvement in prediction accuracy in some competitions.

Semantic segmentation has applications in fields such as automatic driving, robot sensing, medical image analysis, defect detection of infrastructures, etc. In these applications, suitable models are used, depending on the characteristics and amount of available data. Not only the state-of-the-art models but also the pioneering models are still useful and effective in some fields. For example, in concrete crack detection, FCN was used and showed comparable performance to that of state-of-the-art models [7]. The state-of-the-art models for medical image segmentation are variants of U-Net and FCN [8, 9].

This paper addresses a method for improving prediction accuracy of semantic segmentation methods. For this purpose, transfer learning for semantic segmentation networks and conventional data augmentation have been widely used. The effectiveness of the former is significant, but that of the latter is unsatisfactory, as reported in [2, 3]. In this paper, as an effective method for this purpose, we propose a method that uses pre-processing layers based on an autoencoder [10] ahead of a semantic segmentation network.

The outline of our method is as follows:

(1) As pre-processing layers, place the encoder layers of a convolutional autoencoder network ahead of a semantic segmentation network.
(2) Initialize the pre-processing layers using the weights of the pre-trained autoencoder.
(3) Train the entire network for fine-tuning.

The pre-processing layers extract the features of the raw images and feed them to the semantic segmentation network.

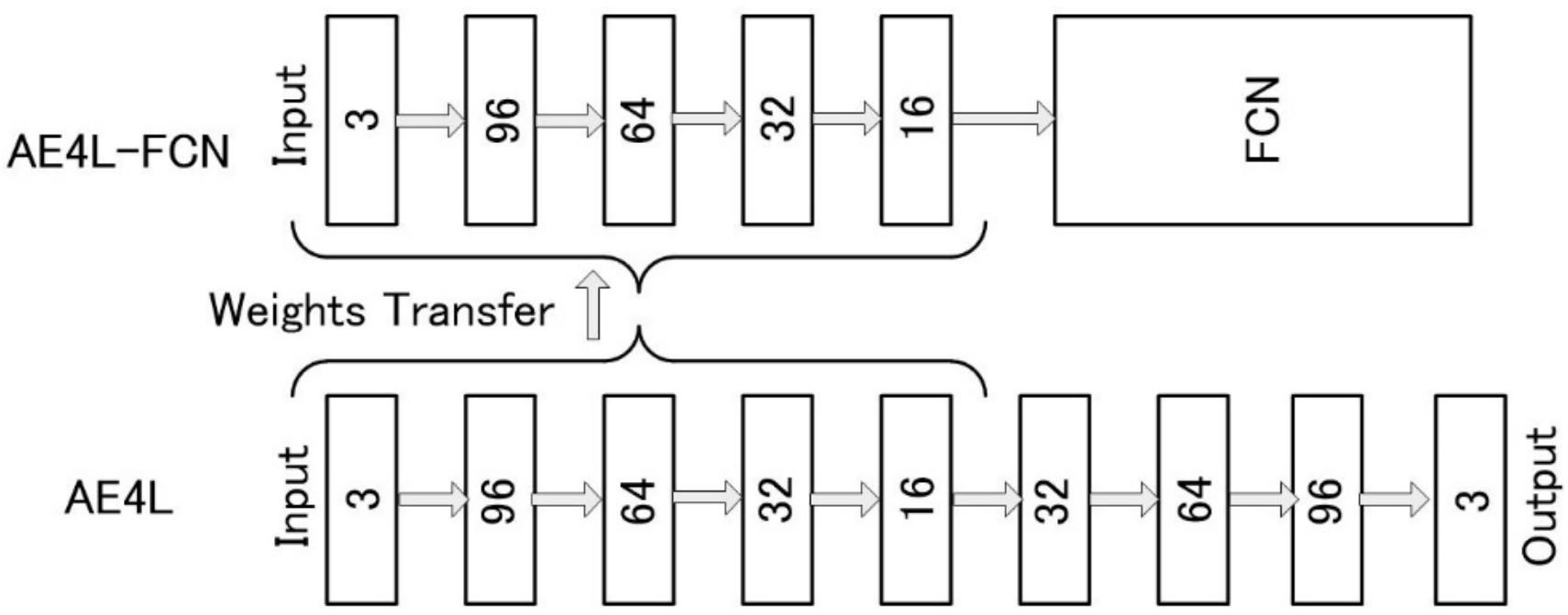

Figure 1: Concept of proposed model. Numbers in figures denote number of filters of each layer.

The concept of the proposed model is illustrated in FIGURE 1. We call this method Convolutional Autoencoder-based Pre-processing Layer (CAEPL) for short. As a target semantic segmentation network, we select FCN. Because FCN is still used in some fields, it is worth doing. We demonstrate the effectiveness of our method through experiments using the cityscapes dataset [11]. In principle, the proposed method can be applied to other convolutional semantic segmentation networks.

The main contributions of our work are summarized as follows:

(1) We propose the CAEPL method which has not yet been published and used in practice.
(2) We design models to realize our method and experimentally prove its effectiveness by applying them to FCN. The Mean IoU of the proposed model with the He normal initialization is 18.66% higher than that of FCN with the He normal initialization.

For semantic segmentation, at present, there is no effective way to improve the prediction accuracy of existing methods. Our method can easily and effectively improve the prediction accuracy of existing semantic segmentation methods. Our method is comparatively simple and the effect on improving prediction accuracy is significant. None have published a method which is the same as or similar to our method and none have used such a method in practice. Therefore, we believe that our method is useful in practice and worthy of being widely known and used by publishing.

The rest of this paper is organized as follows. In Section 2, we describe the research background and related work. In Section 3, we describe the CAEPL method and the models tested in the experiments. In Section 4, we describe the experimental framework, including the training protocol. In Section 5, we present the results of the experiments and discuss them. Finally, in Section 6, we present our conclusions.

## 2. RESEARCH BACKGROUND

In this section, we describe the research background and the motivation behind the proposed method.

An autoencoder is a neural network that can learn to compress and reconstruct input data, such as images, using hidden layers of neurons. The autoencoder model consists of two parts: an encoder and a decoder [10]. The encoder takes the input data x and compresses them into a lower-dimensional representation called the latent space representation or code: $\boldsymbol{h} = f(\boldsymbol{x})$. The decoder then reconstructs the input data from the latent space representation: $\boldsymbol{r} = g(\boldsymbol{h})$. Autoencoders are trained using the backpropagation of the error derivatives. Generally, as the error, the reconstruction error (the difference between the input and output) is used. The reconstruction error is calculated using various loss functions, such as the mean squared error, binary cross-entropy. The loss function used depends on the type of data reconstructed.

There are two types of autoencoders according to the network architecture: a fully connected layer network type [12] and a convolutional layer network type [13]. The feature map of each layer is adjusted by the number of neurons in each layer for the former and the number of filters in each layer for the latter. Compared to the former, the latter are better suited for image processing, because it utilizes the full capability of convolutional neural networks to exploit the image structure.

Among many applications of autoencoders, we focused exclusively on the type in which the codes themselves are used. In [12], Hinton and Salakhutdinov showed that codes produced by autoencoders using a fully connected multilayer neural network are better than those produced by principal component analysis algorithms. Martinez-Murcia et al. [13] used codes produced by convolutional autoencoders to study the manifold structure of an Alzheimer's disease. For video images, Sauvalle et al. [14] proposed a method to compute the foreground/background segmentation masks which uses an autoencoder to learn a low dimensional manifold based on the assumption that the dynamic background of a scene can be modeled by it. Betechuoh [15] used codes produced by a fully connected three-layer autoencoder to classify HIV status. The proposed model yielded an accuracy of 92%, which was better than that of 84% obtained from a conventional feedforward neural network model. Golomb et al. [16] proposed neural networks to identify the human sex. The processing was performed using two separate networks: an image compression network and a sex-identifying network. Both networks were fully connected there-layer networks. The compressed image (code) was inputted into the sex-identifying network. The network's average error rate of 8.1% was better than that of humans 11.6%. Sellami et al. [17] proposed a hyperspectral image classification method which uses, as a part of the input, the fusion of the spectral and spatial features extracted from the hyperspectral image by a multi-view deep autoencoder model. The experimental results showed that the proposed method is competitive in classification performance compared to the state-of the-art methods.

In particular, we noted the good results of Golomb et al. [16] and Sellami et. al. 17]. These results demonstrate that the codes produced by appropriate autoencoders represent the underlying latent fundamental features of the input data. This result is supported by the findings of Hinton and Salakhutdinov [12]. Inspired by these, we hit upon the key idea of our method: inputting the codes produced by an autoencoder into a semantic segmentation model improves its prediction accuracy. This was realized by placing the encoder layers of a convolutional autoencoder network ahead of a semantic segmentation network.

Transfer learning aims at improving the performance of a target learner in the target domain by transferring knowledge contained in different but related source domains [18]. For a deep convolutional neural network, Yosinski et al. [19] experimentally showed that initializing a network with transferred features can improve the generalization performance after fine-tuning on the target dataset. There have been many successful studies on transfer learning, for example, in the training of FCN [2, 3] for PASCAL VOC 2011, the pre-trained VGG16 net was used. Oquab et al. [20] transferred the pre-trained parameters produced in image classification to object or action classification and obtained good results. Observing the good results of these studies, we completed the idea of our method by employing transfer learning: initialize the pre-processing layers using the weights of the pre-trained autoencoder, and train the entire network to fine-tune.

For conventional data augmentation, randomly mirroring and randomly jittering of the input image were used in [2, 3], although they yielded no noticeable improvement. In [5, 6], randomly scaling and randomly mirroring of the input image were used, although the effectiveness of these methods has not been clearly described. Yang et al. [21] reported the Mean IoU with and without data augmentation for several semantic segmentation methods, and the effectiveness of data augmentation.

## 3. CAEPL METHOD AND TEST MODELS

In this section, we describe the CAEPL method and the models tested in the experiments.

### 3.1 CAEPL Method

In Section 1 and Section 2, we described the outline of the CAEPL method. More concretely, we construct and train a model of the CAEPL method as follows (see FIGURE 1).

(1) Design and construct a convolutional autoencoder network (AE4L). Then, train the convolutional autoencoder network.
(2) Construct a model (AE4L-FCN) which has the encoder layers of the convolutional autoencoder network (pre-processing layers) ahead of a semantic segmentation network (FCN). Then, initialize the encoder layers using the weights of the pre-trained autoencoder.
(3) Train the model for fine-tuning.

The pre-processing layers (the encoder layers) extract the underlying latent fundamental features of the raw images and feed them to the semantic segmentation network. Therefore, we need to design the convolutional autoencoder network so that the model can produce good predictions. We experimentally explore such a model.

### 3.2 Convolutional Autoencoder

As explained in Section 2, an autoencoder model consists of two parts: an encoder and a decoder [10]. The encoder takes the input data x and compresses them into a lower-dimensional representation called the latent space representation or code: $\boldsymbol{h} = f(\boldsymbol{x})$. The decoder then reconstructs the input data from the latent space representation: $\boldsymbol{r} = g(\boldsymbol{h})$. The training process minimizes the following loss function (reconstruction error): $L\left(\boldsymbol{x}, g\left(f(\boldsymbol{x})\right)\right)$. Bengio et al. [22] showed that corrupting input images by adding noise and training to denoise them using the reconstruction error is an effective method for implicitly capturing the underlying data-generating distribution. In this case, since the input $\boldsymbol{x}$ is corrupted into $\tilde{\boldsymbol{x}}$, the loss function is $L\left(\boldsymbol{x}, g\left(f(\tilde{\boldsymbol{x}})\right)\right)$.

The autoencoder layers must be symmetrical across the code layer. The layers are designed so that features can be extracted according to the purpose of the model. In our method, the code extracted by the encoder is the input to FCN that takes three channels (R, G, and B). We used an autoencoder of the convolutional layer network type. We used neither pooling nor uppooling in the network, because pooling discards the useful image details that are essential for extracting the features of the images. That is, all the layers have the same size as that of the input image. The feature maps are produced according to the number of filters in each layer. Referring to the models presented by Hinton and Salakhutdinov [12], we designed models using four or three layers with different filter numbers.

We use a filter with a size of 3×3 for all the convolutional layers. The convolution stride is one pixel, and the padding is one pixel. We use a batch normalization layer after each convolutional layer.

### 3.3 FCN

According to the notation in [2, 3], we denote the three models as FCN-32s, FCN-16s, and FCN-8s, where the numbers represent the upsampling strides of the finest feature map. We use FCN-8s in our experiments and refers to FCN-8s as FCN. We use a batch normalization layer after each convolutional layer.

### 3.4 Test Models

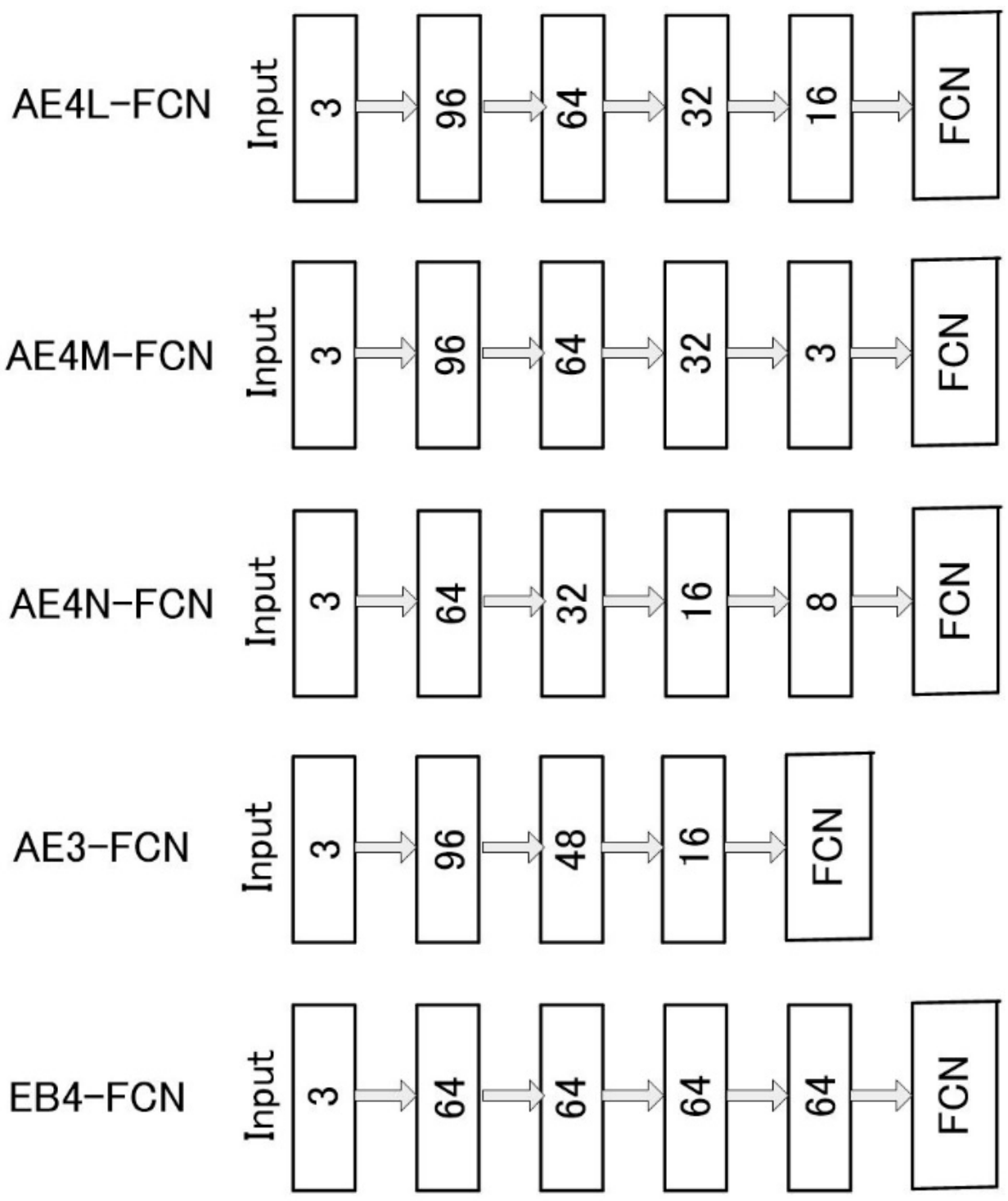

Figure 2: Test models. Numbers in figures denote number of filters of each layer.

The test models are illustrated in FIGURE 1 and FIGURE 2: AE4L-FCN, AE4M-FCN, AE4N-FCN, AE3-FCN, and FB4-FCN. AE4L represents the autoencoder with a four-layer encoder and decoder, and AE4L-FCN represents the model with the encoder layers of AE4L ahead of FCN. Similarly, AE4M-FCN, AE4N-FCN, and AE3-FCN represent the models with the encoder layers of the corresponding autoencoders, that is, AE4M, AE4N, and AE3, respectively, ahead of FCN. EB4 represents four layers that are not autoencoders, and EB4-FCN represents the model with EB4 ahead of FCN. The target model is AE4L-FCN, and the other models are used to compare the performance. In these models, we use a batch normalization layer after each convolutional layer.

### 3.5 Implementation of Models

We implemented the models using Keras in TensorFlow with Python and Jupyter Notebooks. We used an NVIDIA Quadro GV100 GPU.

## 4. EXPERIMENTAL FRAMEWORK

In this section, we describe the experimental framework including the training protocol.

### 4.1 Tasks and Metrics

We performed the experiments to examine the prediction accuracy of the models described in Section 3. In the experiment, we used Cityscapes dataset, which was used to evaluate many models. For evaluation, we used the pixel-wise accuracy (Pix. Acc.) and the mean of class-wise intersection-over-union (standard Mean IoU). In either case, pixels with void labels do not contribute to the score.

### 4.2 Cityscapes Dataset

The cityscapes dataset [11] is a dataset used for semantic urban scene understanding. It contains 5,000 high-quality traffic-related images collected from 50 cities in different seasons, which are finely annotated at a pixel level. The annotation consists of 19 classes belonging to eight categories of foreground objects, background, and overall scene layout. The image size is 2048×1024. The images are divided into sets of 2,975, 500, and 1,525 for training, validation, and testing, respectively. In addition, coarsely annotated 20,000 images are provided. The annotations for the test images are publicly withheld for benchmarking purposes.

### 4.3 Previous Studies on Training of FCN

Elaborate and tedious training procedures were required to achieve the best performance of FCN [2, 3].

(1) Long et al. experimented with both staged and all-at-once trainings. In the staged version, they trained the single-stream FCN-32s, then upgraded to the two-stream FCN-16s and continued learning, and finally upgraded to the three-stream FCN-8s and finished learning. At each stage, the network was initialized with the parameters of the earlier network and trained end-to-end. The learning rate of each stage was set to $1.0^{-2}$ times that of the earlier stage. In the all-at-once training, only FCN-8s were trained end-to-end. To adjust the disparate feature scale, they scaled each stream using a fixed constant that was selected to approximately equalize the average feature norms across streams.

(2) In the training of FCN for PASCAL VOC 2011 [2, 3], they used the pre-trained VGG16 net. They indicated that training from scratch resulted in a substantially lower accuracy.

Some papers reported the prediction accuracy on the cityscapes dataset. In [11], Cordts et al. adopted VGG16 net [1] and utilized the PASCAL context setup [2, 3]. Each image was split into two halves, with a sufficiently large overlap. They first trained on the training set (train) until the performance on the validation set (val) saturated and then retrained on train+val with the same number of epochs. As a result, they achieved a Mean IoU of 65.3% on the test set. In [23], Guist used images downscaled by a factor of two. He initialized FCN-32s with the VGG16 net weights and performed the staged training on the training set and achieved a Mean IoU of 60.8% on the validation set. We can see that initializing the network using the VGG16 net weights and the staged training contribute to the higher prediction accuracy, and the more training data, the higher the prediction accuracy.

## 4.4 Strategy for Experiments

The purpose of our experiments is not to achieve the best performance of FCN. It is to examine and show the relative differences in the prediction accuracy between the proposed models and FCN. Accordingly, we decided on the basic strategy for training FCN and the test models as follows:

(1) We use the all-at-once training, because the staged training is tedious and time-consuming. We use batch normalization instead of the aforementioned fixed constant to scale each stream.

(2) Mostly, we use training from scratch with a He normal network initializer [24]. For the layers corresponding to the VGG16 layer net, we test the cases initializing them by the weights of the pre-trained VGG16 net [25] for FCN and some test models described later for comparison purposes.

(3) We use only the training (train) dataset in the training stage and use the validation (val) dataset for validation.

(4) The input image is downscaled by a factor of two, that is, 1024×512, owing to memory restrictions.

## 4.5 Training Protocol of Autoencoder

**Input image** According to [22], we corrupt the input images by adding salt-and-pepper noise (probability 0.5 of corrupting each bit, setting it to 1 or 0 with pixel probability 0.5).

**Loss function** As explained in Section 3.1, the loss function (reconstruction error) of the autoencoder model is $L\left(\boldsymbol{x}, g\left(f(\tilde{\boldsymbol{x}})\right)\right)$. Because the input image has three channels (R, G, and B), the theoretically correct loss function is the mean squared error. For comparison purposes, we tested both the mean squared error and binary-cross-entropy loss for the target model AE4L-FCN and compared the Mean IoU of the prediction results: 0.4477 for the former and 0.4503 for the latter. Because our goal is to achieve a higher prediction accuracy of the test models, we decided to use the binary-cross-entropy loss.

**Activation function** The activation functions of all convolutional layers are ReLu [26], except for the last layer. That of the last layer is the modified hyperbolic tangent: o.5tanh($\boldsymbol{x}$)+0.5.

**Hyperparameters** We performed preliminary experiments to explore the optimum values of the hyperparameters using AE4L and decided on them, considering the computation time and necessary memory size. We adopted stochastic gradient descent (SGD) with Nesterov momentum: the learning rate is $1.0e^{-4}$ and the momentum is 0.9. The batch size is 4. As the kernel regularizer, we use an L2-norm of $1.0e^{-3}$.

**Evaluation** We monitor the minimum of val_loss using keras.callbacks.ModelCheckpoint and save the weights of the model at the epoch of a minimum of val_loss. We stop the training at the epoch of 2000, when the minimum of val_loss does not improve.

## 4.6 Training Protocol of FCN

**Activation function** The activation functions of all convolutional layers are ReLu [26], except for the last layer. The last layer is the softmax layer.

**Hyperparameters** We performed preliminary experiments using FCN to explore the optimum values of the hyperparameters and decided on them, considering the computation time and necessary memory size. We adopted SGD with Nesterov momentum: the learning rate is $1.0e^{-4}$ and the momentum is 0.9. The batch size is 5. As the kernel regularizer, we use an L2-norm of $1.0e^{-4}$.

**Evaluation** We monitor the maximum of val_acc using keras.callbacks.ModelCheckpoint and save the weights of the model at the epoch of a maximum of val_acc. We stop training at 3000 epochs when the maximum val_acc does not improve. The best segmented images are produced using the weights of the model at the epoch of the maximum val_acc.

### 4.7 Training Protocol of Test Models

**Activation function** The activation functions of all convolutional layers are ReLu [26], except for the last layer. The last layer is the softmax layer.

**Hyperparameters** We performed preliminary experiments on AE4L-FCN to explore the optimum values of the hyperparameters and decided on them, considering the computation time and necessary memory size. We adopted SGD with Nesterov momentum: the learning rate is $1.0e^{-4}$ and the momentum is 0.9. The batch size is 5. As the kernel regularizer, we use an L2-norm of $1.0e^{-3}$ for the layers of the encoder and that of $5.0e^{-4}$ for those of FCN.

**Evaluation** We monitor the maximum of val_acc using keras.callbacks.ModelCheckpoint and save the weights of the model at the epoch of a maximum of val_acc. We stop training at 3000 epochs when the maximum of val_acc does not improve. The best segmented images are produced using the weights of the model at the epoch of the maximum val_acc.

## 5. RESULTS AND DISCUSSION

In this section, we present and analyze the results of the experiments and discuss them.

### 5.1 Results

#### 5.1.1 AE4L

In FIGURE 3, we show the loss during the training and validation of AE4L. In FIGURE 4, we show examples of the original images (left), corrupted images (center), and predicted images (right) for AE4L. It should be noted that the predicted images are very similar to the original images, and so the original images are well reconstructed by AE4L.

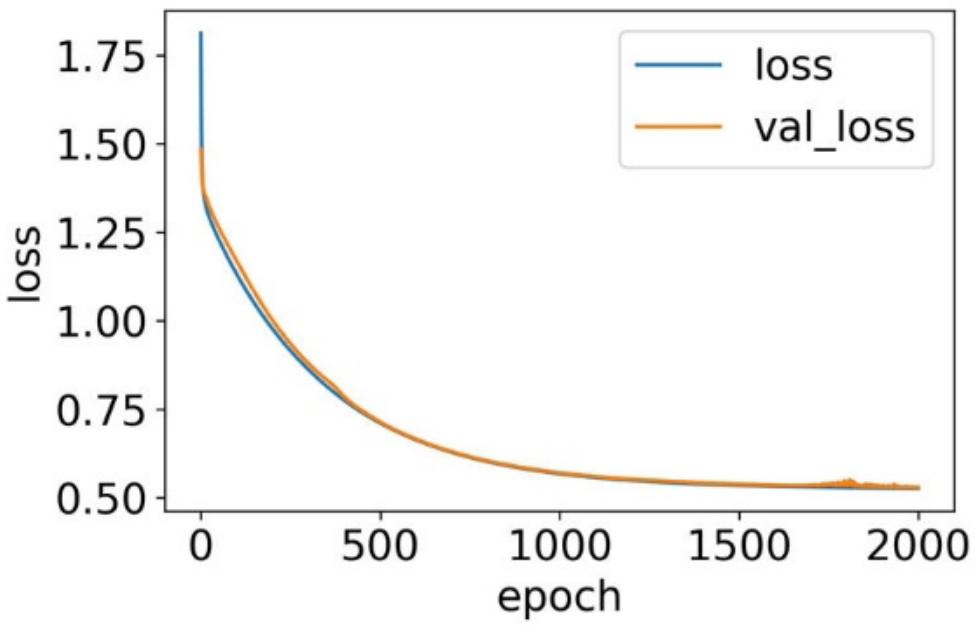

Figure 3: Loss during training and validation for AE4L.

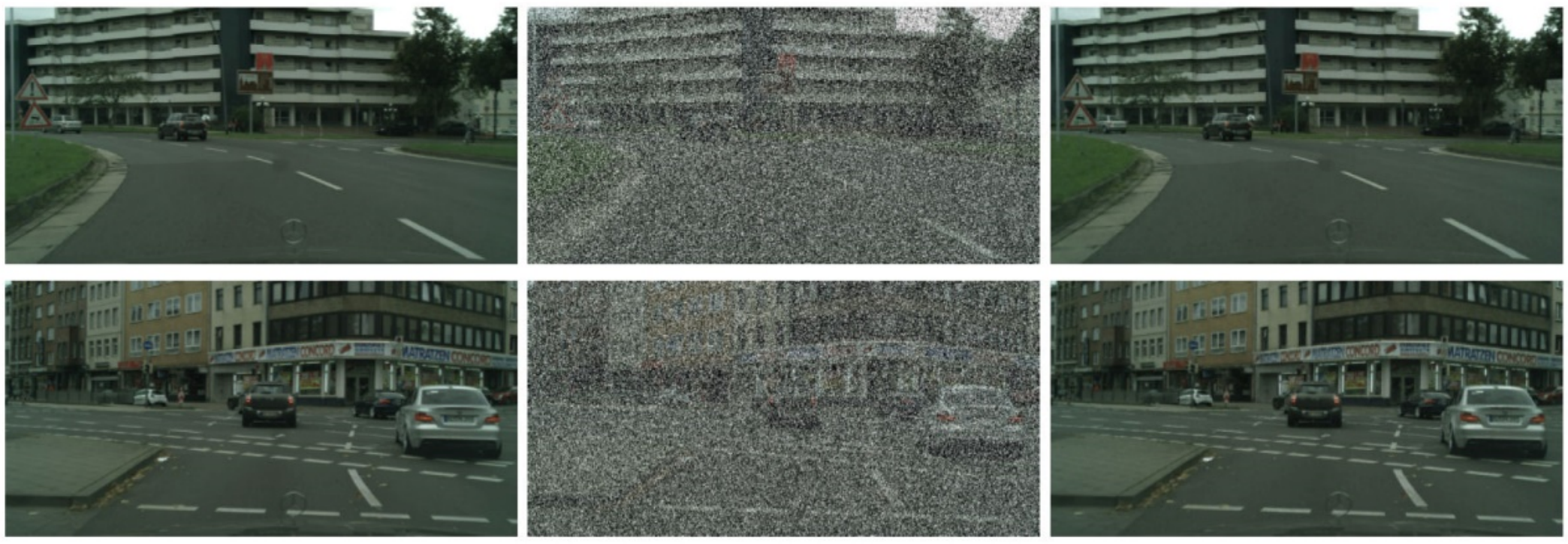

Figure 4: Examples of original image (left), corrupted image (center), and predicted image (right) for AE4L.

### 5.1.2 FCN and AE4L-FCN

Here, we show the results of FCN and AE4L-FCN and compare them.

In FIGURE 5 and FIGURE 6, we illustrate the accuracy (left) and loss (right) during the training and validation of FCN and AE4L-FCN, respectively.

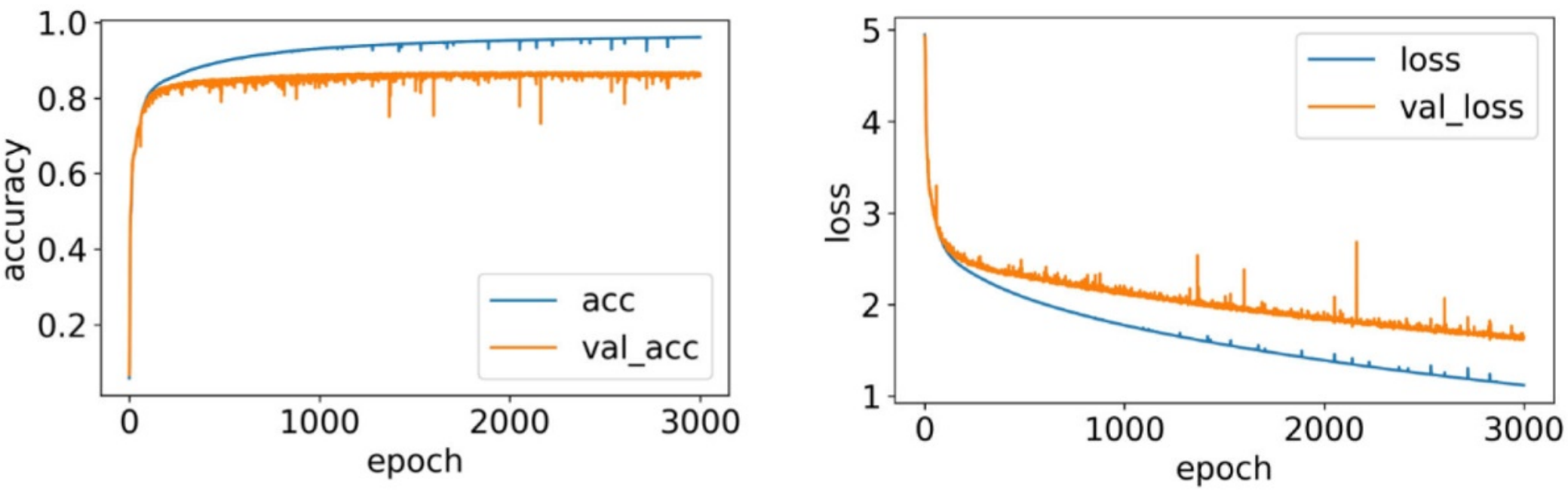

Figure 5: Accuracy (left) and loss (right) during training and validation for FCN.

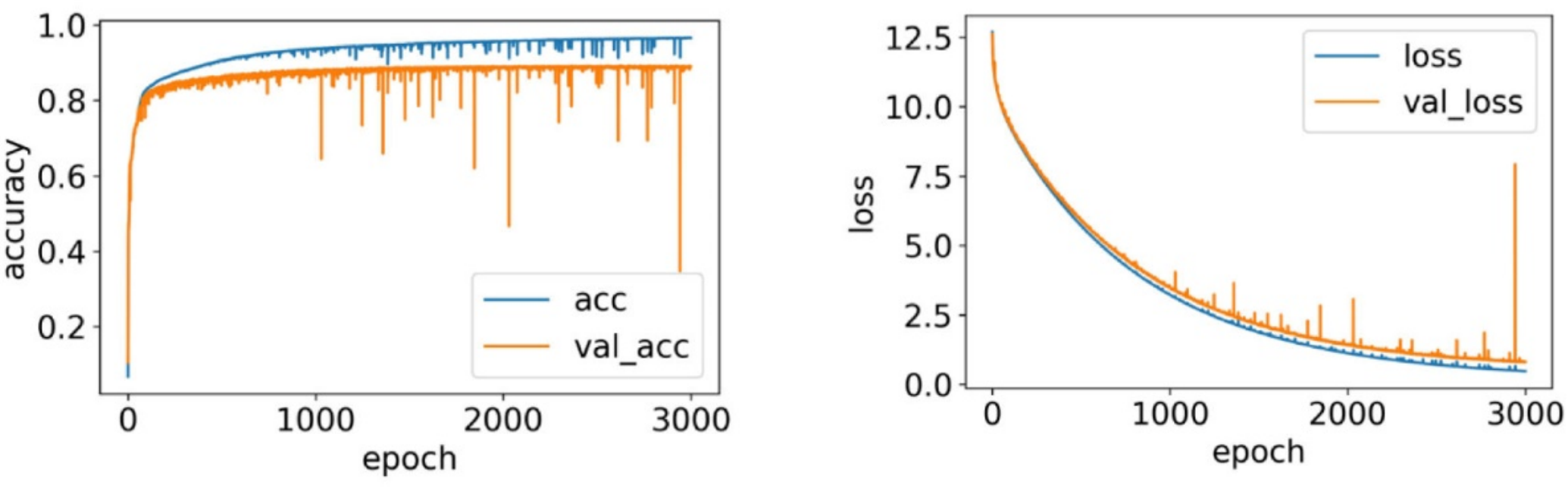

Figure 6: Accuracy (left) and loss (right) during training and validation for AE4L-FCN.

In TABLE 1, we show the prediction accuracy of FCN. The Mean IoU of FCN with the He normal initialization was 37.95%. In TABLE 2, we show the prediction accuracy of the test models. The Mean IoU of the target model AE4L-FCN was 45.03%. It is 18.66% higher than that of FCN with the He normal initialization, and it is 6.03% higher than that of FCN with the VGG16 net weights initialization. The second row shows the case with the He normal initialization instead of the pre-trained weights for the encoder layers of AE4L-FCN. The Mean IoU in this case was 43.69%. It should be noted that AE4L-FCN achieves a significantly higher Mean IoU compared to that of FCN with the He normal initialization, and this holds true even when the pre-trained weights are not used for the encoder layers of AE4L-FCN.

Table 1: Prediction accuracy of FCN on cityscapes dataset. Models were trained on train dataset and evaluated using val dataset.

| Parameter Initializer | Mean IoU | Pix. Acc. |
|---|---|---|
| he-normal | 37.95 | 87.21 |
| VGG16 | 42.47 | 88.60 |

Table 2: Prediction accuracy of test models on cityscapes dataset. Encod. Weights denotes initializing pre-processing layers using weights of pre-trained autoencoder. Models were trained on train dataset and evaluated using val dataset.

| Model | Encoder Weights | Mean IoU | Pix. Acc. |
|---|---|---|---|
| **AE4L-FCN** | **use** | **45.03** | **89.04** |
| AE4L-FCN | not use | 43.69 | 88.73 |
| AE4M-FCN | use | 43.94 | 88.84 |
| AE4N-FCN | use | 43.82 | 88.71 |
| AE3-FCN | use | 44.23 | 88.91 |
| EB4-FCN | not use | 42.12 | 87.77 |

In FIGURE 7, we show the best (first row) and worst (second row) predictions of FCN and the predictions of AE4L-FCN for the same input image. Correspondingly, in TABLE 3**,** we show the best and worst Mean IoU's of FCN and the Mean IoU's of AE4L-FCN for the same input image.

In FIGURE 8, we show the best (first row) and worst (second row) predictions of AE4L-FCN and the predictions of FCN for the same input image. Correspondingly, in TABLE4, we show the best and worst Mean IoU's of AE4L-FCN and the Mean IoU's of FCN for the same input image. Both the best and worst Mean IoU's of AE4L-FCN are higher than those of FCN. In the best predicted segmentation of AE4L-FCN, a small person in the original image was recognized.

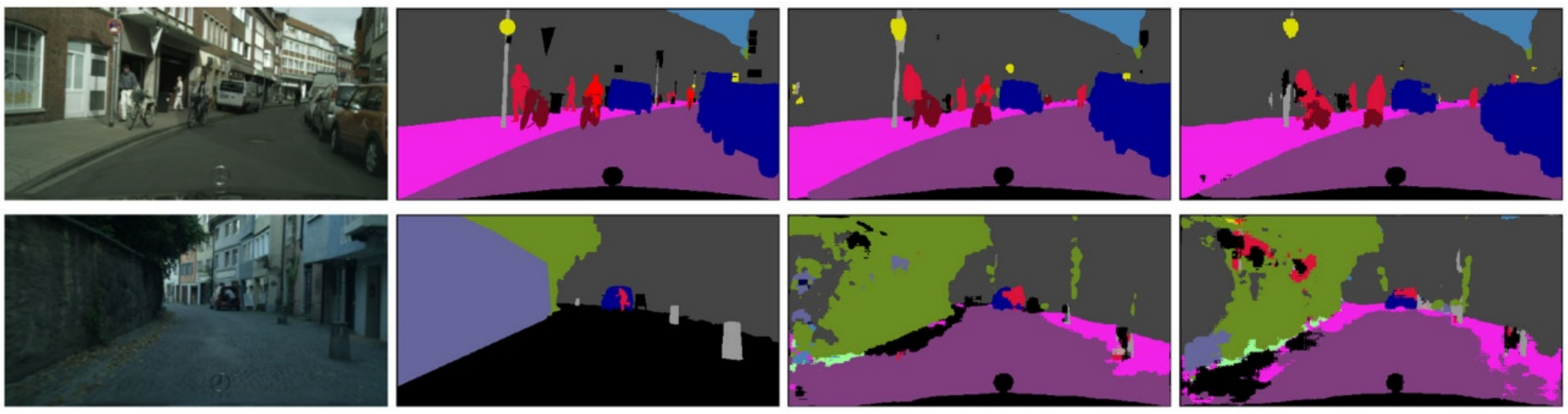

Figure 7: Best (first row) and worst (second row) predictions of FCN and predictions of AE4L-FCN for the same input image: input image (first column), ground truth image (second column), predicted segmentation (third column) of FCN, and predicted segmentation of AE4L-FCN for the same input image (last column).

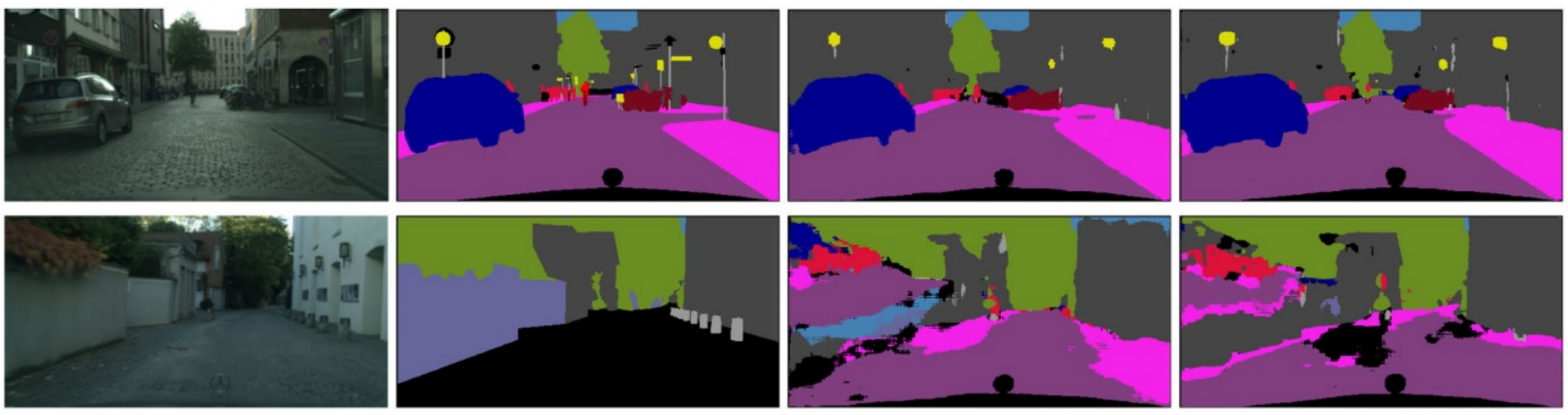

Figure 8: Best (first row) and worst (second row) predictions of AE4L-FCN and predictions of FCN for the same input image: input image (first column), ground truth image (second column), predicted segmentation (third column) of AE4L-FCN, and predicted segmentation of FCN for the same input image (last column).

Table 3: Best and worst Mean IoU's of FCN and Mean IoU's of AE4L-FCN for the same input image.

| B. or W. | FCN | AE4L-FCN |
|---|---|---|
| Best | 0.6312 | 0.6290 |
| Worst | 0.1055 | 0.1435 |

Table 4: Best and worst Mean IoU's of ACE4L-FCN and Mean IoU's of FCN for the same input image. the input image.

| B. or W. | AE4L-FCN | FCN |
|---|---|---|
| Best | 0.7042 | 0.5842 |
| Worst | 0.1423 | 0.1252 |

In TABLE 5, we show the number of parameters of FCN, AE4L-FCN, and AE4L. The total number of parameters for AE4L is very small. The number of total parameters of AE4L-FCN is 0.07% larger than that of FCN. This difference is very small.

Table 5: Number of parameters.

| Model | Trainable | Non-trainable | Total |
|---|---|---|---|
| FCN | 134,473,244 | 160 | 134,473,404 |
| AE4L-FCN | 134,567,020 | 704 | 134,567,724 |
| AE4L | 163,059 | 800 | 163,859 |

In TABLE 6, we show the computation time (days) for the training of FCN, AE4L-FCN, and AE4L. The computation time of AE4L-FCN is 1.48 times that of FCN. The total computation time of AE4L-FCN and AE4L is 1.94 times that of FCN. Although the number of trainable parameters for AE4L is very small, its computation time is long.

Table 6: Computation time (days) for training.

| FCN | AE4L-FCN | AE4L | AE4L-FCN+ AE4L |
|---|---|---|---|
| 18.93 | 27.97 | 8.69 | 36.66 |

### 5.1.3 Other test models

Here, we analyze the results of the other test models.

(1) AE4M-FCN is a modified model of AE4L-FCN that has a code with three filters for the encoder. The Mean IoU of it was 43.94%.

(2) AE4N-FCN is a modified model of AE4L-FCN with a smaller number of filters for each layer than AE4L-FCN. The Mean IoU of it was 43.82%.

(3) AE3-FCN is a modified model of AE4L-FCN with the encoder layers of AE3L. The Mean IoU of it was 44.23%.

(4) The Mean IoU's of these modified models of AE4L-FCN are slightly lower than that of AE4L-FCN. However, it should be noted that the Mean IoU's of these modified models are significantly higher than that of FCN with the He normal initialization.

(5) EB4-FCN is a model with four layers of 64 filters without pre-training. The Mean IoU of it was 42.12%. It is significantly higher than that of FCN with the He normal initialization. It should be noted that placing a simple network such as this ahead of FCN results in a significant improvement in prediction accuracy.

### 5.1.4 Effect of VGG16 net weights transferring

For FCN, the Mean IoU of the case with the VGG16 net weights initialization [25] was 42.47% (see TABLE 1), which is 11.91% higher than that of the case with the He normal initialization. This effect is significant.

In TABLE 7, we show the prediction accuracy of AE4L-FCN and AE4M-FCN with the weights transferring to the FCN layers. The first row shows the case in which the first layer of FCN used the He normal initialization, and the remaining layers used the initialization with the corresponding weights of the VGG16 net layers. The Mean IoU of it was 45.39%. It is slightly higher than that of AE4L-FCN with the He normal initialization. The second row shows the case of AE4M-FCN with the VGG16 net weights initialization. The mean IoU of it was 44.14%. It is slightly higher than that of the case with the He normal initialization (see TABLE 2). For AE4L-FCN and AE4M-FCN, the effect of the VGG16 net weights transferring is small.

Table 7: Prediction accuracy of AE4L-FCN and AE4M-FCN with VGG16 net weights transferring to FCN layers on cityscapes dataset. Models were trained on train dataset and evaluated using val dataset.

| Model | Weights Initialization | Mean IoU | Pix. Acc. |
|---|---|---|---|
| AE4L-FCN | He normal & VGG15 | 45.39 | 89.18 |
| AE4M-FCN | VGG16 | 44.14 | 88.75 |

## 5.2 Summary of Results and Discussion

(1) The Mean IoU of the target model WE4L-FCN with the He normal initialization is 18.66% higher than that of FCN with the He normal initialization, and it is 6.03% higher than that of FCN with the VGG16 net weights initialization. In addition, those of the modified models of AE4L-FCN are significantly higher than that of FCN with the He normal initialization. These results provide strong evidence that the proposed method is significantly effective in improving the prediction accuracy of FCN. It should be noted that the proposed method is comparatively simple, whereas the effect on improving prediction accuracy is significant.

(2) Comparing the differences between loss and val-loss, and those between acc and val-acc, we see that those of WE4L-FCN are much smaller than those of FCN (see FIGURE 5 and FIGURE 6). This is the clear evidence that the proposed method significantly improves the generalization ability of FCN.
(3) The increase in the number of parameters by applying the proposed method to FCN is very small. However, that of the computation time is substantially large. This increase in the computation time is a shortcoming of the proposed method.
(4) For AE4L-FCN and AE4M-FCN, the effect of the VGG16 weights transferring is small. This is a weak point of the proposed method.
(5) Yang et al. [21] reported the effects of the conventional data augmentation on the PASCAL VOC dataset. According to the results in Table 2, the Mean IoU's were 0.53% and 0.95% for DeepLavV3+ and PSPNet, respectively, higher than the case without the dada augmentation. Although the dataset and segmentation method used are different from ours, these results provide indirect evidence that our method is much superior to the conventional data augmentation.
(6) The Mean IoU's of the state-of-the-art models on the cityscapes dataset were reported as follows. For DeepLabv3 [5] and PSPNet [6] using both the fine and coarse datasets, they were 81.3 and 80.2, respectively. Since these models are more elaborate than FCN and trained using extra coarse dataset, they naturally produce much higher prediction accuracy than the proposed model. However, as described in introduction, the pioneering models are still useful and effective in some fields. For example, in concrete crack detection, FCN was used and showed comparable performance to that of state-of-the-art models [7]. The state-of-the-art models for medical image segmentation are variants of U-Net and FCN [8, 9]. Therefore, we believe that improving their prediction accuracy is significant and important.
(7) In principle, the proposed method can be applied to other semantic segmentation methods, such as U-net [4] and PSPNet [6]. As age does not allow us to try it, we hope that the interested readers will try it and report good results.
(8) For semantic segmentation, at present, there is no effective way to improve the prediction accuracy of existing methods. Our method can easily and effectively improve the prediction accuracy of existing semantic segmentation methods. Our method is comparatively simple and the effect on improving prediction accuracy is significant. None have published a method which is the same as or similar to our method and none have used such a method in practice. Therefore, we believe that our method is useful in practice and worthy of being widely known and used by publishing.

## 6. CONCLUSIONS

In this paper, to improve prediction accuracy of semantic segmentation methods, we proposed the CAEPL method as follows: (1) construct a neural network that has pre-processing layers based on a convolutional autoencoder ahead of a semantic segmentation network, and (2) train the entire network initialized by the weights of the pre-trained autoencoder. We applied this method to FCN and experimentally examined its prediction accuracy on the cityscapes dataset. The Mean IoU of the proposed target model with the He normal initialization is 18.66% higher than that of FCN with the He normal initialization. In addition, those of the modified models of the target model are significantly higher than that of FCN with the He normal initialization. The accuracy and loss curves during the training showed that these are resulting from the improvement of the generalization ability. All of these results provide strong evidence that the proposed method is significantly effective in improving the prediction accuracy of FCN. The proposed method has the following features: it is comparatively simple and the effect on improving the generalization ability and prediction accuracy of FCN is significant; the increase in the number of parameters by using it is very small, and that in the computation time is substantially large. In principle, the proposed method can be applied to other semantic segmentation methods. We believe that our method is useful in practice and worthy of being widely known and used.

**Reproducibility**
The main source codes used in the experiments are available at https://github.com/dplneuralnet/

**Declaration of competing interest**
This research was performed on a personal basis and did not receive any specific funding.
The author declares that no competing interests exist.

**Hisashi Shimodaira** was born in 1941 in Japan. He received the BEng, MEng, and DEng degrees from Tokyo Metropolitan University in 1969, 1971, and 1982, respectively. He joined Bunkyo University in Japan in1996 and retired in 2011. He is currently a free researcher and devoting the rest of his life to achieving his research goal of publishing an excellent paper. His research interests include computer vision and artificial intelligence, especially deep learning neural network. The current focus of his research is semantic segmentation method.